# Tamil Language Computing: the Present and the Future

**Kengatharaiyer Sarveswaran**

Department of Computer Science
University of Jaffna, Sri Lanka.

sarves@univ.jfn.ac.lk

## Abstract

This paper delves into the text processing aspects of Language Computing, which enables computers to understand, interpret, and generate human language. Focusing on tasks such as speech recognition, machine translation, sentiment analysis, text summarization, and language modelling, language computing integrates disciplines including linguistics, computer science, and cognitive psychology to create meaningful human-computer interactions. Recent advancements in deep learning have made computers more accessible and capable of independent learning and adaptation. In examining the landscape of language computing, the paper emphasises foundational work like encoding, where Tamil transitioned from ASCII to Unicode, enhancing digital communication. It discusses the development of computational resources, including raw data, dictionaries, glossaries, annotated data, and computational grammars, necessary for effective language processing. The challenges of linguistic annotation, the creation of treebanks, and the training of large language models are also covered, emphasising the need for high-quality, annotated data and advanced language models. The paper underscores the importance of building practical applications for languages like Tamil to address everyday communication needs, highlighting gaps in current technology. It calls for increased research collaboration, digitization of historical texts, and fostering digital usage to ensure the comprehensive development of Tamil language processing, ultimately enhancing global communication and access to digital services.[1]

## 1.0 Introduction

Language computing, also known as Natural Language Processing (NLP), is a field of Artificial Intelligence that focuses on the interaction between computers and human languages, primarily in the form of speech and text. It involves developing algorithms and systems that enable computers to understand, interpret, and generate human language in a way that is both meaningful and useful. This encompasses a wide range of tasks, including speech recognition, machine translation, sentiment analysis, text summarisation, and language modelling.

Although nowadays mostly computer scientists and computational linguists are working on language computing, Language computing is an interdisciplinary area which combines elements of linguistics, computer science, cognitive psychology, and many more disciplines to create tools, applications, and services that can be useful for humans in various ways, including tools and services like virtual assistants, automated customer service, information retrieval, and language education.

---

[1] This is the write-up of the address delivered at the 29th Annual Sessions of the Jaffna Science Association, held from March 29-31, 2023, at the University of Jaffna. The slides from this talk can be found:
https://www.slideshare.net/slideshow/tamil-language-computing-the-present-and-the-future/269413540
ChatGPT-4 was used to polish (not to write) the article.

The goal of language computing is to make computers or machines to process, and understand human languages and make meaningful inferences in order to make computers more accessible for human needs.

## 1.2 Why Language Computing

Language computing has become an integral part of our daily lives, simplifying various aspects of communication and information processing. From virtual assistants[2] to machine translation services[3] and aiding differently abled people[4], language computing is everywhere, breaking down language and communication barriers that once challenged global communication. By understanding and producing human language, these technologies make processes more efficient and faster, allowing real-time communication and access to information in different languages and cultures. Language computing not only improves personal and professional interactions but also democratises access to knowledge and digital services.

The business opportunities of language computing are enormous, driving innovation in a variety of industries. Companies use these technologies to analyse large amounts of text data, creating knowledge and insights and obtaining information to inform decision-making. Automated customer service, sentiment analysis, and marketing are just a few examples of how language computing is changing industries. By making business processes more efficient, language computing helps companies save time and resources, ultimately leading to greater productivity and profitability.[5]

Beyond its commercial applications, language computing supports research and scholarship in the fields of language, linguistics, and the humanities. Researchers around the world use natural language processing tools to study linguistic patterns, compare linguistic structures, explore the evolution of languages, and understand cultural nuances. This technological support is invaluable for academic research, allowing scholars to analyse large data sets that would otherwise be unmanageable.

## 1.2 Human ↔ Machine communication

Until very recently, computers needed to be trained in specific ways using structured information and precise instructions. This meant that humans had to communicate with computers in a highly structured manner, adhering to strict syntax and formats (Programming languages) to ensure that the machines could understand and execute the commands, but computers could generate languages that humans could understand (shown in Figure 01).

---

[2] https://assistant.google.com/
[3] https://translate.google.com
[4] https://www.seeingai.com/
[5] https://mck.co/3sED0NU

However, today, instructions for computers can be fed using natural language, making interactions more intuitive and accessible, and making computers accessible by everyone.

In the recent deep learning era, computers have gained the capability to learn independently. By utilising vast amounts of data from various sources of human input, they can construct their own knowledge maps and improve their understanding and performance over time (Figure 02). This self-learning ability allows computers to adapt to new information, refine their algorithms, and enhance their functionality without needing explicit programming for every new task. Consequently, this shift not only simplifies human-computer interaction but also opens up new possibilities for automation and innovation across numerous fields.

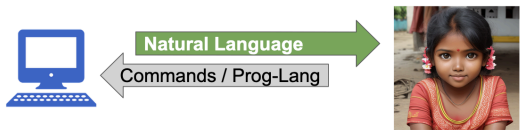
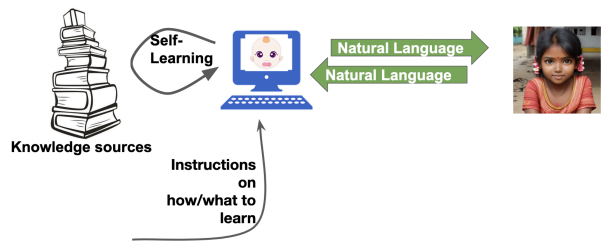

Figure 01: Traditional Computer-Human communication method

Figure 02: Present Computer-Human communication method

## 2.0 Language computing landscape

The language computing landscape can be visualised as shown in Figure 03. The foundational work of language computing enables the processing of any language. Building computational resources provides the necessary power for machines to process languages effectively. Based on these two components, language applications can then be developed. This section will examine the current status of each of these aspects in detail. Most of these computational resources are primarily built by computational linguists who have expertise in both Linguistics and Computer Science.

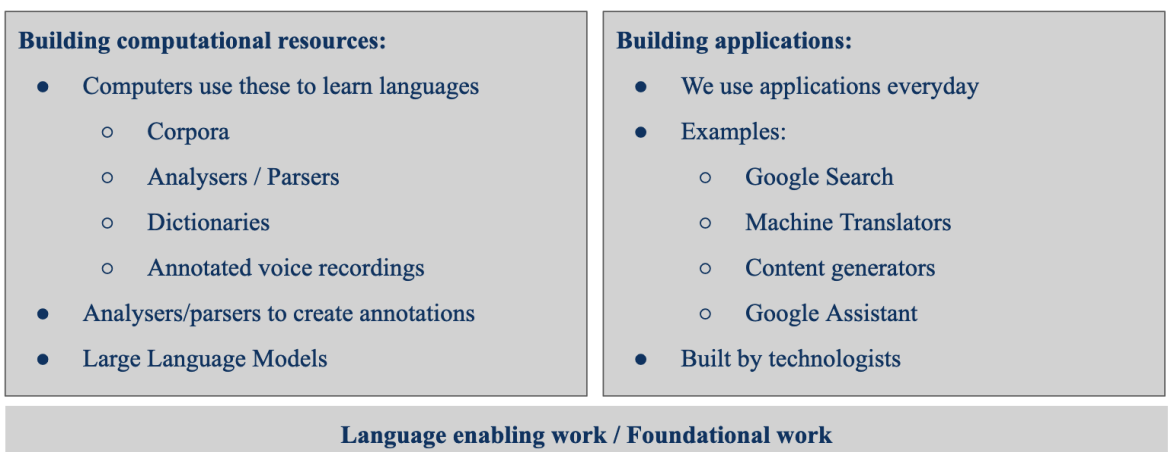

Figure 03: Language computing landscape

## 2.1 Language enabling/foundational work

Encoding refers to the mapping in which the characters we write are stored in a computer in the form of numbers. In the early days, Tamil was written using American Standard Code for Information Interchange (ASCII) encoding. However, various individuals modified the font rules to create numerous fonts that could display Tamil letters using ASCII encoding. The problem with this approach was that the exact font used for typing had to be applied to view the text correctly. Otherwise, the characters would not be displayed properly because the letters were actually stored as Latin characters. They appeared in Tamil only when the correct font rules were used.

This approach presented a huge problem. Since computers natively do not understand Tamil, it was challenging to develop applications for Tamil as we do for English. For instance, tasks such as sorting text, publishing content on the internet, or sending emails in Tamil were either impossible or very cumbersome. The lack of a standardised encoding system meant that Tamil text could not be easily shared or processed, significantly limiting its use in digital communication and content creation.

Figure 04: Tamil Unicode Chart

Later, in 1993, Tamil was included in Unicode version 1.1 encoding. Unicode aims to encode all the living scripts in the world so that computers can understand each script natively. Initially, a 128 code-block was allocated for Tamil, and Tamil characters were encoded within this space.[6] Instead of encoding each character individually, Tamil vowels, consonants (consonants with a modifier), vowel modifiers, numbers, and some symbols were encoded. These were initially sufficient to write basic Tamil characters using Unicode. However, since the standardisation was done following other Indo-Aryan scripts, specifically the Devanagari script, issues like sorting exist (ல, வ, ழ, ள, ற, ன are not encoded in the correct order as used in Tamil) in the current code chart.

However, Unicode has enabled computers to understand Tamil characters natively. In addition to regular Tamil letters, special characters like numbers and symbols were also included in Unicode. Recently, Tamil fractions have been added to Unicode, which will help computers understand classical texts in the future.

Several fonts have been created using Tamil Unicode, allowing old character styles to be easily reproduced using Unicode font rules. Despite these advancements, the printing sector still struggles to adapt to Unicode, as some applications do not support Tamil well and sometimes break Tamil characters.

Moreover, awareness about Unicode is lacking. Although many people write in Unicode using mobile keyboards, some still use the old ASCII-based approach to type in Tamil.

---

[6] https://www.unicode.org/charts/PDF/U0B80.pdf

Additionally, some institutions and universities require their members to write articles in the ASCII-based approach. There are several scripts available to convert text from ASCII to Unicode. However, without knowing the correct font used for typing, it would be difficult to convert the text to Unicode accurately. While the foundational support for Tamil exists, there is still a long way to go to fully implement and practise Unicode standards.

## 2.2 Building computational resources

As shown in Figure 03, there are several computational resources available to feed language and linguistic knowledge to computers. This section will discuss some of the key resources used in language computing.

### 2.2.1 Raw data

There is a large amount of raw Tamil data available from various sources, including Wikipedia, news websites, and reports published by institutions. Recently, there have been efforts to compile these resources for training computers in natural language processing. For instance, Common Crawl[7] is a popular resource widely used for training machine learning models. This compilation of web data provides a vast dataset that can be leveraged to improve the capabilities of language models in understanding and generating Tamil text.

However, existing efforts do not adequately catalogue these data sources in a meaningful way. Tamil varies significantly among regions, including differences in dialects and glossaries. When building an application or system for a particular region, it is essential to use data specific to that region. Currently, we lack properly compiled data sources that reflect these regional variations. Therefore, more work is needed to organise and catalogue Tamil data sources to ensure they are useful for region-specific applications and systems.

### 2.2.2 Dictionaries, Glossaries, Word Lists, and Gazetteers

We also lack comprehensive online dictionaries. While there are sources like Wikidata, the University of Madras lexicon,[8] and various glossaries available online,[9] these resources are not sufficient for capturing contemporary language. More importantly, glossaries need to be regularly updated to accommodate new terms and evolving language usage. This ongoing updating process is crucial to ensure that the resources remain relevant and useful for modern applications.

Given that Tamils are now living all over the world, it is also important to create and maintain up-to-date gazetteers. Gazetteers provide valuable geographic information, and having an updated gazetteer will support a wide range of applications, from academic research to practical tools for navigation and location-based services. Maintaining current and accurate geographic data is essential for effectively serving the global Tamil community and addressing their specific needs.

---

[7] https://commoncrawl.org/
[8] https://dsal.uchicago.edu/dictionaries/tamil-lex/
[9] https://www.language.lk/en/resources/terminology/

## 2.2.3 Annotated data

Annotations enrich data by adding additional layers of information, making it more valuable for analysis and processing. There are several levels of annotations that can be added. Annotations can be applied at the document level, and then further detailed at the content level, such as the paragraph level, sentence level, word level, morpheme level, and phoneme level. Annotations done at the document level are widely referred to as metadata annotations. The other annotations, which include those at the paragraph, sentence, word, morpheme, and phoneme levels, are mostly considered linguistic annotations. These linguistic annotations provide detailed information on the structure and meaning of the text, aiding in various applications such as natural language processing, linguistic research, and machine learning.

Understanding linguistic structure involves multiple layers of analysis. Phonology studies how phonemes make up sounds. Morphology examines how morphemes form words. Syntax explores how words combine to form sentences. Semantics investigates how language conveys meaning, while pragmatics looks at how language is used in context. By incorporating these aspects into annotations, data becomes richer and more useful for comprehensive linguistic analysis and advanced computational applications.

வந்தான் வா|+verb|+fin|+sim|+strong|+past=(ந்)த்|+3sgm=ஆன்

Figure 05: Morphological analysis of the Tamil verb "வந்தான்" translates to "Came.he"

Although some linguistic annotations have been made at the levels of part of speech, morphology, and syntax, there is still a long way to go. Figure 05 shows an example for inflectional morphological analysis. Even in morphology, only inflectional morphology is handled in Tamil, where the analysis is limited to given words of the same part of speech. This means that while basic morphological analysis can be performed, more complex aspects of morphology, such as derivational processes, remain underdeveloped.

The biggest challenge we face in linguistic annotation is the lack of skilled annotators. These annotations cannot be done by native speakers alone; they require special training to understand linguistic structures. Furthermore, the process is extremely time-consuming and tedious. This combination of limited expertise and the labour-intensive nature of the task significantly hampers progress of the linguistic annotation process.

## 2.2.4 Computational Grammars

Computational grammars are essential resources that enable computers to understand language structure. These grammars can be written using linguistic formalisms such as Lexical Functional Grammar (LFG) and Head-driven Phrase Structure Grammar (HPSG). Writing computational grammars requires deep linguistic knowledge and an understanding of these formalisms. Additionally, expertise in modelling them using computers is crucial. There have been attempts to build computational grammars for Tamil using LFG. An analysis is shown in Figure 06. As illustrated in the figure, the analysis provides two types of structures: constituency structures and dependency structures. So far, these grammars handle simple sentences, and the work is ongoing.

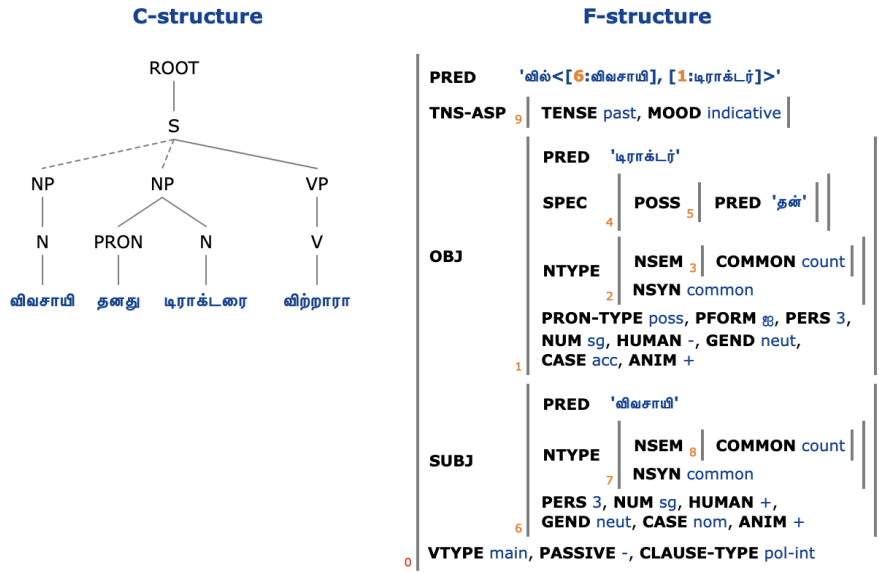

Figure 06: An analysis of a Tamil sentence using the LFG Formalism

### 2.2.5 Treebanks

Treebanks are valuable linguistic resources for training machines with rich linguistic information. They typically include annotations such as lemmas, parts of speech, morphology, and syntax. Dependency treebanks are particularly suitable for languages like Tamil, which have a free word order. Universal Dependencies (UD)[10] is a widely used formalism for building treebanks, covering over 160 languages to date. Since UD was developed with cross-lingual analysis in mind, treebanks created using this formalism can easily be compared with those of other languages, facilitating comparative studies.

There have also been efforts to develop treebanks using computational grammars. To date, two Tamil treebanks have been published in the Universal Dependencies repository. These are relatively small treebanks, but there are ongoing efforts to build a larger treebank with 100,000 tokens, which may soon be available in the repository.

### 2.2.6 Language models

Another key computational resource in natural language processing is the Large Language Model (LLM),[11] which is based on deep learning and neural networks. These models are trained on extensive datasets that include nearly everything written on the internet or available in digital form. Large language models, such as GPT and BERT, are capable of recognising, summarising, translating, predicting, and generating text and other content. Some of these models are multilingual, allowing them to handle multiple languages and facilitate cross-linguistic tasks. The vast amounts of data they are trained on enable them to understand and produce human-like text with remarkable accuracy.

Training these models involves various algorithms and different types of neural network architectures. For example, encoder models convert given text to numerical representations that machines can understand, encoder-decoder models transform text from one form to another, and decoder models convert numerical representations back into text. These models can be customised for a variety of tasks, such as translation, summarisation, and text generation. For instance, Fairseq can be customised to perform translation tasks. The flexibility and adaptability of these models, combined with their deep learning foundations, make them invaluable tools for a wide range of applications in natural language processing and beyond.

---

[10] https://universaldependencies.org/
[11] https://developers.google.com/machine-learning/resources

Building advanced language models such as GPT-3 presents several significant challenges. These models require an enormous amount of high-quality data for effective training. Additionally, the computational resources needed are substantial; for instance, training GPT-3 reportedly costs around $4 million, and running ChatGPT costs approximately $100,000 per day.[12] The training processes for these models are often not transparent, making it difficult for external parties to understand or replicate the methods used. Due to these factors, developing and maintaining such models is typically beyond the reach of small organisations and educational institutions, particularly in parts of the world with limited resources.

**2.3 Building applications**

There are many applications being created for Tamil, including spell checkers, grammar checkers, text-to-speech applications, and machine translation applications. Some notable examples include Tamil spell checkers like Vaani,[13] and Tamil text-to-speech[14] applications such as Google Text-to-Speech. However, unlike the broad spectrum of applications available for other resource-rich languages like English, there are not as many applications addressing all the problems faced by Tamil speakers.

Most of these applications currently solve high-level problems, such as aiding in academic research and professional settings, but there are few, if any, applications that can be used by grassroots people to address their everyday issues. For example, while there are advanced tools for machine translation and educational purposes, there is a lack of practical, user-friendly applications for everyday tasks like casual communication, local business transactions, and accessibility tools for the elderly. This gap highlights the need for more development in this area to make technology more accessible and useful for Tamil speakers in their daily lives.

## 3.0 Future of Tamil Language Processing

The future of Tamil language processing appears to be centred on Large Language Models (LLMs), which are revolutionising the way we work and retrieve knowledge. Globally, there is significant investment in LLMs, leading to the development of applications like ChatGPT. Therefore, it is crucial that we also focus on leveraging these models for the Tamil language to ensure quality service and beneficial outcomes. While this is a primary focus, we must not neglect the study of the Tamil language itself, necessitating substantial investment in Tamil linguistic studies.

---

[12] https://www.cnbc.com/2023/03/13/chatgpt-and-generative-ai-are-booming-but-at-a-very-expensive-price.html
[13] http://vaani.neechalkaran.com/
[14] https://ttsfree.com/text-to-speech/tamil-india

## 3.1 Focus on Quality Data

Although a significant amount of Tamil data exists, it is not as extensive as for some other languages and is often poorly organised and annotated. For example, resources like Common Crawl and corpora from AiPharat contain valuable data but suffer from duplication and lack of structure. Given Tamil's linguistic diversity, including regional dialects and the distinction between spoken and written forms, it is essential to organise existing data and annotate it with metadata indicating type and dialect. Additionally, we must continue collecting more information to build a comprehensive, high-quality dataset.

## 3.2 Focus on Digitising More Tamil Data

Tamil boasts a continuous literary tradition spanning millennia, with much of its old literature still on ola leafs and in printed books that have yet to be digitised. This vast body of knowledge remains untapped by current models. Developing technologies to digitise these texts is crucial. Furthermore, since the language has evolved over time, the structure and vocabulary of older texts differ from contemporary usage. We need strategies to ensure that models do not get confused by these variations.

## 3.3 Focus on Annotated Data

We also need more annotated data to evaluate and fine-tune large language models accurately. Detailed annotations capturing the nuances of the Tamil language are necessary to assess model performance and ensure context-appropriate output. Additionally, annotating data to capture cultural specificities is essential to prevent models from generating inappropriate content. Tamil's rich and diverse culture must be reflected in these annotations. Investing in computational linguistics studies and expanding related courses will cultivate a future generation of trained computational linguists capable of fine-tuning, evaluating, and building applications.

## 3.4 Focus on Research Collaboration

Enhancing research collaboration is vital. While there are global efforts related to Tamil technology, institutional collaboration is often weak, possibly due to funding constraints. Few institutions and grant organisations are willing to invest in Tamil-specific research, making it difficult for funded projects to involve more collaborators. Securing more funding and encouraging collaborative projects can address this issue. For instance, partnering with others to apply for substantial grants from various global organisations could be a viable strategy.

**3.5 Focus on Increased Tamil Usage**

Increasing the usage of Tamil in daily tasks and across devices will encourage major corporations like Google and Facebook to develop resources for the language. Writing more in Tamil, building content, and engaging with Tamil websites and ads can drive this initiative. It is also crucial to motivate the Tamil diaspora to use the language more frequently. Evidence from Latin America shows that local language usage can significantly boost sales, suggesting similar potential for Tamil.

# 4.0 Conclusion

Language computing is a rapidly evolving interdisciplinary field that integrates disciplines including linguistics, computer science, and cognitive psychology to enable computers to process and understand human languages. Language computing has transformed daily life by facilitating real-time communication, breaking down language barriers, and democratising access to digital services across different languages and cultures. Despite significant progress, challenges remain in developing resources for low resource languages like Tamil. While the transition from ASCII to Unicode has improved digital communication, further efforts are needed to standardise and raise awareness of Unicode. Comprehensive computational resources, including updated dictionaries, glossaries, and annotated data, are essential to support advanced NLP applications.

The advancement of Tamil language processing is at a critical juncture, driven by the emergence of large language models (LLMs). Although these technologies have revolutionised communication and information access, Tamil language computing still faces challenges related to data quality, linguistic diversity, and digital representation. To fully harness the potential of NLP for Tamil, future efforts should focus on collecting and curating high-quality data, digitising historical texts, and providing detailed linguistic annotations. Research collaboration and increased digital usage of Tamil are crucial. By leveraging LLM capabilities and fostering a collaborative environment, Tamil language computing can keep pace with global advancements while preserving and promoting Tamil's linguistic and cultural heritage, ultimately enriching the lives of Tamil speakers worldwide and driving innovation in various fields.